\documentclass[twocolumn]{article}

\usepackage{arxiv}
\usepackage{amsmath}
\usepackage{algorithm}
\usepackage{algpseudocode}
\usepackage[utf8]{inputenc} 
\usepackage[T1]{fontenc}    
\usepackage{hyperref}       
\usepackage{url}            
\usepackage{float}
\usepackage{subfigure}
\usepackage{caption}
\usepackage{graphicx}
\usepackage{newtxtext}
\usepackage{newtxmath}
\usepackage{longtable}
\usepackage{multirow} 
\usepackage{booktabs} 
\usepackage{tabularx,ragged2e}
\usepackage{booktabs}       
\usepackage{amsfonts}       
\usepackage{nicefrac}       
\usepackage{microtype}      
\usepackage{cleveref}       
\usepackage{lipsum}         
\usepackage{graphicx}
\usepackage{natbib}
\usepackage{doi}
\usepackage{abstract}
\usepackage{subcaption}
\title{Fuzzy hyperparameters update in a second order optimization}

\newif\ifuniqueAffiliation
\uniqueAffiliationtrue

\ifuniqueAffiliation 
\author{ \href{a.bensadok@logicwizard.dz}{\includegraphics[scale=0.06]{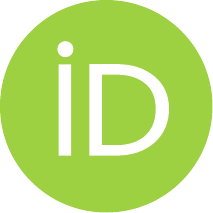}\hspace{1mm}Abdelaziz.~Bensadok} \\
	CTO \& Lead Researcher\\
	LogicWizard EURL\\
	Djelfa, Algeria \\
	\texttt{a.bensadok@logicwizard.dz} \\
	\And
	\href{M.Z.Babar@leeds.ac.uk}{\includegraphics[scale=0.06]{orcid.pdf}\hspace{1mm}Muhammad Z.~Babar} \\
	School of Computing\\
	University of Leeds\\
	Leeds, UK \\
	\texttt{M.Z.Babar@leeds.ac.uk} \\
}
\else

\usepackage{authblk}

\newbox{\orcid}\sbox{\orcid}{\includegraphics[scale=0.06]{orcid.pdf}} 
\author[1]{%
	\href{a.bensadok@logicwizard.dz}{\usebox{\orcid}\hspace{1mm}Abdelaziz ~Bensadok\thanks{\texttt{a.bensadok@logicwizard.dz}}}%
}
\author[1,2]{%
	\href{M.Z.Babar@leeds.ac.uk}{\usebox{\orcid}\hspace{1mm}Muhammad Z .~Babar\thanks{\texttt{M.Z.Babar@leeds.ac.uk}}}%
}
\affil[1]{CTO \& Lead Researcher, Logicwizard EURL, Djelfa, Algeria}
\affil[2]{School of Computing, University of Leeds, Leeds, UK}
\fi

\renewcommand{\shorttitle}{Fuzzy Hyperparameters Update in a Second Order Optimization}

\hypersetup{
pdftitle={Fuzzy Hyperparameters Update in a Second Order Optimization},
pdfsubject={cs.AI},
pdfauthor={Abdelaziz.~Bensadok},
pdfkeywords={Optimization, Fuzzy Logic, Inverse Hessian Matrix, Learning rate, Momentum},
}

\begin{document}
	
\twocolumn[
\maketitle
]

\begin{abstract}
	\textbf{\
	This research will present a hybrid approach to accelerate convergence in a second order optimization. An online finite difference  approximation of the diagonal Hessian matrix will be introduced, along with fuzzy inferencing of several hyperparameters. Competitive results have been achieved
	}
\end{abstract}

\section{Introduction}
Deep learning models have become increasingly popular in recent years, with their success attributed to gradient-based optimization techniques such as stochastic gradient descent (SGD) and its variants. However, these first-order methods can be limited in their ability to efficiently navigate complex optimization landscapes. Second-order optimization methods, which leverage information about the curvature of loss function, have emerged as a promising alternative \citep{Sun2019}. This research focuses on the use of finite difference approximation of Hessian matrix, along with fuzzy tuning of key hyperparameters. Initial results have shown substantial improvement in accuracy compared to state-of-the-art optimization methods. In view of this, the project will tackle the optimization problem from a rarely visited angle: Fuzzy Logic.  First, an online approximation of the Hessian matrix is introduced and tested, then a fuzzy logic system to tune the  hyperparameters is implemented , this hybrid approach may decrease the computational burden, and competitive results can be delivered.

\section{Second Order Optimization Review}

Hessian-based optimization methods use second-order derivatives to estimate the curvature of the loss function and offer efficient optimization. yet they tend to be computationally expensive, particularly for high-dimensional problems like deep learning. To address this issue, several techniques have been developed, each with its own advantages and drawbacks.

Conjugate gradient (CG) methods \citep{Hestenes1952} bridge the gap between first-order and second-order gradient approaches. By generating a series of conjugated directions based on the gradient at known points, they allow for faster convergence than the steepest descent method. However, CG methods as noted by \citet{Sun2019} involve more complex calculations compared to first-order gradient techniques.

In contrast, Newton's method leverages second-order gradient information from the Hessian matrix for even faster convergence. Although it achieves quadratic convergence under specific conditions \citep{Bonnans2006}, it requires considerable computing time and substantial storage space to calculate and store the Hessian matrix's inverse matrix at each iteration as noticed by \citet{Sun2019}.

To reduce computational time, quasi-Newton methods approximate the Hessian matrix or its inverse matrix using an estimated matrix \citep{Schoenberg2001}. \citet{Sun2019} explain that methods such as DFP, BFGS, and LBFGS can achieve super linear convergence but require significant storage space, limiting their suitability for large-scale optimization problems.

An alternative approach is the use of stochastic quasi-Newton methods, which integrate stochastic optimization strategies. Examples like online-LBFGS \citep{Zhu1997} can effectively handle large-scale machine learning problems, but its calculation is more complex than the stochastic gradient method as emphasized by \citet{Sun2019}.

The Hessian Free (HF) method \citep{Martens2011} on the other hand addresses high-dimensional optimization by using the conjugate gradient for sub-optimization, avoiding the costly computation of the inverse Hessian matrix. While well-suited for high-dimensional optimization, \citet{Sun2019} note that the computational cost of the matrix-vector product in the HF method increases linearly with the growth of training data.

Finally, the sub-sampled Hessian Free method \citep{Kohler2017}, is a method tackles large-scale machine learning optimization problems by employing a stochastic gradient and sub-sampled Hessian-vector during the update process. Although it offers a solution for such challenges, \citet{Sun2019} noted that it requires more complex calculations and increased computing time per iteration compared to the stochastic gradient method.

In summary, a variety of techniques developed to overcome the computational challenges associated with Hessian-based optimization methods, each with its own strengths and limitations.

\section{Diagonal Hessian Approximation}

The Hessian matrix is a crucial component in second-order optimization methods, as it provides information about the local curvature of the objective function. However, as mentioned earlier, computing and storing the full Hessian matrix can be computationally expensive, especially for high-dimensional problems. The diagonal Hessian approximation is a technique that addresses this issue by only considering the diagonal elements of the Hessian matrix.

The need for efficient optimization methods has become increasingly important with the growing size and complexity of machine learning models. Full Hessian-based methods, such as Newton's method, necessitate calculating and inverting the Hessian matrix, which, according to \citet{Zhang2017}, results in a computational complexity of $O(n^3)$ for an n-dimensional problem. Conversely, \citet{Jarry2021} suggested that the diagonal Hessian approximation reduces this complexity by considering only the diagonal elements of the Hessian matrix. This approach results in an $O(n)$ complexity while still preserving crucial information about the curvature of the objective function.

Many techniques have been developed to approximate diagonal the Hessian matrix to be used in the second-order optimization methods. Finite differences are one of the simplest methods. It can be employed to compute the diagonal elements of the approximated Hessian. In the optimization process, the finite differences method helps to calculate online the necessary partial derivatives to construct the diagonal Hessian matrix, which is then used to optimize the objective function in high-dimensional problems. The finite differences harness the information on the first derivatives on discrete data points along with the rate of change of a function to estimate the second-order derivatives when exact second-order derivatives are unavailable, difficult to compute, or too computationally expensive. There are two main types of finite differences:

\begin{itemize}
	\item Forward differences: Forward differences approximate the derivative of a function at a specific point using the difference between the function's value at that point and its value at a neighboring point in the positive direction of the independent variable. The formula for the forward difference approximation of the first derivative is:
	\begin{center}
		\begin{equation}
				f'(x) \approx \frac{f(x + h) - f(x)}{h}
		\end{equation}
	\end{center}
	where h is a small positive increment. 
	
	\item Central differences: Central differences approximate the derivative of a function at a specific point using the average difference between the function's values at neighboring points around that point. Central differences provide a more accurate approximation of the derivative than forward differences (Pelinovsky, D. Lectures .n.d). The formula for the central difference approximation of the first derivative is: 
	
	\begin{center}
		\begin{equation}
		f'(x) \approx \frac{f(x + \frac{h}{2}) - f(x - \frac{h}{2})}{h}
		\end{equation}
	\end{center}
	
	where h is a small positive increment. 
\end{itemize}

Throughout the years, many methodologies of diagonal Hessian approximation have been proposed by researchers. However, almost none of these approaches have been adopted in large-scale applications, primarily due to scalability concerns. Prominent applications such as BART and ChatGPT have opted to utilize the ADAM optimizer first introduced by \citet{Kingma2014}, which has become the default optimizer for a myriad of other applications \citep{Yu2020}, owing to its optimal balance between accuracy and computational cost. In recent years, several promising diagonal Hessian approximations have emerged, offering a competitive equilibrium between accuracy and computational expense. Two recently developed approximations that address the limitations of previous methods are Neculai's method \citep{Neculai2020} and Yao et al.'s method \citep{Yao2021}.

\subsection{ Diagonal Approximation of the Hessian by Finite Differences for Unconstrained Optimization}
In this method, \citep{Neculai2020} approximates the diagonal elements of the matrix by combining the classical variational technique with the minimization of the trace of the matrix subject to the weak secant equation, where the diagonal elements are determined by forward or by central finite differences. The search direction is a direction of sufficient descent. The algorithm is equipped with an acceleration scheme and the convergence of the algorithm is linear. Our method is similar to Andrei’s method, but it differs from it in the choice of the step to calculate the second order derivative by finite difference method. While Neculai's method depends on the Wolfe line search conditions to choose the learning rate, our method depends on the Adam algorithm with Exponential Moving Averages of Squared Gradients -which is considered a second-order moment (variance) of the gradients- that is exponentially decaying. Besides that incorporating bias correction for both the first and second moments ensure that the initial estimates for these moments are accurate. Bias correction accounts for the initial zero values of the moments, preventing them from being too low in the early stages of optimization and they are applied in the same way they are applied in Adam.

\subsection{ADAHESSIAN }
\citep{Yao2021} introduced ADAHESSIAN, a novel stochastic optimization algorithm that incorporates approximate curvature information from the loss function with low computational overhead. This algorithm features a fast Hutchinson-based method, spatial averaging, and an RMS exponential moving average. Extensive tests on NLP, CV, and recommendation system tasks show that ADAHESSIAN achieves state-of-the-art results, outperforming other popular optimization methods. The cost per iteration is comparable to first-order methods, and ADAHESSIAN demonstrates improved robustness towards hyperparameter variations.

\section{Fuzzy Logic Based Scheduling: Literature Review}

A Fuzzy Expert System can handle various input types, including vague, distorted, or imprecise data, making them a robust choice when precise inputs are not available. Fuzzy Expert Systems (FES) are tools that manage input uncertainty \citep{Zadeh1983}. In the field of learning rate optimization, there have been limited attempts to utilize FES for deep neural network hyperparameter optimization, primarily due to the availability of many schedulers that offer minimal computational burden compared to FES. These schedulers include Step Decay, Exponential Decay, Time-based Decay, Cosine Annealing, Cyclic Learning Rate (CLR), One Cycle Policy, ReduceLROnPlateau, Warm-up, Warm Restarts, and more.

\citet{Shaout2000} were among the first to a fuzzy logic modification system to enhance the standard backpropagation algorithm by dynamically adjusting the learning rate parameter using a fuzzy system controller. This approach, based on problem domain heuristics, aims to accelerate training time and improve the stability of convergence. The efficacy of the proposed system was demonstrated through its application in character recognition tasks.

\citet{Wen2019} developed a fuzzy method for adjusting learning rates in multilayer neural networks (MNNs), which increased learning accuracy by 2\% - 3\% compared to a constant learning rate. This method demonstrated robustness and insensitivity to parameter changes due to approximate reasoning, and can be applied to more complex tasks in convolutional neural networks.

Recently \citet{Hosseinzadeh2022} proposed a Fuzzy Scheduling implementation for learning rate scheduling. When using the same initial random weights and learning rates, a convolutional neural network trained with the fuzzy scheduler achieved higher accuracy than other scheduling methods (Cosine Restart Decay and Exponential Decay).

\section{Diagonal Hessian Approximation: Introducing Our Method}

In a deep learning optimization, the Hessian matrix, which is the second-order partial derivative of the objective function with respect to the weights and biases, plays a crucial role in determining the convergence of optimization algorithms. Computing the full Hessian can be expensive, especially for high-dimensional problems. Hence, approximations are often used to reduce the computational cost. The main equation that relates the hessian and the learning rate with the first derivative the loss function is as follows:

\begin{equation}
	H \Delta W = g
\end{equation}

So the rate of the change in the weights can be simply deduced:

\begin{equation}
	\Delta W = H^{-1} \cdot g
\end{equation}

Where \(H\) is the Hessian, \(\Delta W\) is the incremental change in weight, and \(g\) is the gradient of the loss function with respect to the weight \(w\). A detailed form of this equation is as follows:

\begin{equation}
	\begin{bmatrix}
		\frac{\partial^2 L}{\partial w_1^2} & \frac{\partial^2 L}{\partial w_1 \partial w_2} & \cdots & \frac{\partial^2 L}{\partial w_1 \partial w_n} \\
		\frac{\partial^2 L}{\partial w_2 \partial w_1} & \frac{\partial^2 L}{\partial w_2^2} & \cdots & \frac{\partial^2 L}{\partial w_2 \partial w_n} \\
		\vdots & \vdots & \ddots & \vdots \\
		\frac{\partial^2 L}{\partial w_n \partial w_1} & \frac{\partial^2 L}{\partial w_n \partial w_2} & \cdots & \frac{\partial^2 L}{\partial w_n^2}
	\end{bmatrix}
	\begin{bmatrix}
		\Delta w_1 \\
		\Delta w_2 \\
		\vdots \\
		\Delta w_n
	\end{bmatrix}
	=
	\begin{bmatrix}
		\frac{\partial L}{\partial w_1} \\
		\frac{\partial L}{\partial w_2} \\
		\vdots \\
		\frac{\partial L}{\partial w_n}
	\end{bmatrix}.
\end{equation}

A diagonal approximation of the Hessian matrix means that only the diagonal elements of the matrix are computed, while the off-diagonal elements are assumed to be zero. This approximation is cheaper to compute than the full Hessian and can be particularly useful when the off-diagonal elements are relatively small.

\begin{equation}
	\begin{bmatrix}
		\frac{\partial^2 L}{\partial w_1^2} & 0 & \cdots & 0 \\
		0 & \frac{\partial^2 L}{\partial w_2^2} & \cdots & 0 \\
		\vdots & \vdots & \ddots & \vdots \\
		0 & 0 & \cdots & \frac{\partial^2 L}{\partial w_n^2}
	\end{bmatrix}
	\begin{bmatrix}
		\Delta w_1 \\
		\Delta w_2 \\
		\vdots \\
		\Delta w_n
	\end{bmatrix}
	=
	\begin{bmatrix}
		\frac{\partial L}{\partial w_1} \\
		\frac{\partial L}{\partial w_2} \\
		\vdots \\
		\frac{\partial L}{\partial w_n}
	\end{bmatrix}.
\end{equation}

One way to compute the diagonal approximation of the Hessian is by using finite differences. The idea is to advance each weight in the model by a small amount (usually it is a small learning rate step ) in the gradient descent algorithm. Then, evaluate the loss function's first-order partial derivatives at the two points the starting point and the second point reached by stepping one small learning rate in gradient descent. 

calculate the first derivatives with back propagation then evaluate the difference of values of the first-order partial derivatives at the original point, and the reached point dividing by the stepping amount, an estimate of the second-order partial derivatives (the diagonal elements of the Hessian) can be obtained.

The quality of the approximation will depend on the specific problem and the accuracy of the finite differences. In some cases, the diagonal approximation may not capture the true curvature of the function, which could lead to slow convergence or incorrect results for this, a smaller learning rate can better approximate the second derivative.

Consider a deep learning optimization problem, where the goal is to minimize a loss function $L(w)$, with $w$ being a vector of $n$ weights $(w = [w_1, w_2, \dots, w_n])$. The Hessian matrix, $H$, is an $n$ by $n$ matrix of second-order partial derivatives:
\vspace{1em}

\begin{center}
	\begin{equation}
		H_{ij} = \frac{\partial^2 L(w)}{\partial w_i \partial w_j}, \text{ for } i,j=1,2,\dots,n
	\end{equation}
\end{center}
\vspace{1em}

The full Hessian matrix can be expensive to compute (equation \ref{eq:der1}), especially for large-scale problems. Therefore, we often use approximations to reduce computational costs. One such approximation is the diagonal Hessian matrix, where only the diagonal elements are computed, and the off-diagonal elements are set to zero as in equation \ref{eq:der2} :
\vspace{1em}

\begin{equation}
	\mathbf{H}_D = \begin{bmatrix}
		\frac{\partial L}{\partial w_1} & 0 & \cdots & 0 \\
		0 & \frac{\partial L}{\partial w_2} & \cdots & 0 \\
		\vdots & \vdots & \ddots & \vdots \\
		0 & 0 & \cdots & \frac{\partial L}{\partial w_n}
	\end{bmatrix}.
	\label{eq:der2}
\end{equation}
\vspace{1em}

To compute the diagonal Hessian matrix using finite differences, we can apply the following procedure:

\begin{itemize}
	\item Select a small perturbation value, r (in deep learning it is a small learning rate around $r \sim 10^{-6}$ so we can calculate the second derivatives online.
	\item For each variable $w_i \hspace{0.1cm} (i = 1, 2, \dots, n)$, compute the first-order partial derivatives of loss function $L(w)$ with respect to the weight $wi$ at two points: $w$ and $w + r$, where r is the learning rate along the ith coordinate axis by back propagation: \begin{align}
		\frac{\partial L(w)}{\partial w_i} \quad \text{and} \quad \frac{\partial L(w+r)}{\partial w_i}
	\end{align}
	
	\item Estimate the second-order partial derivatives (diagonal elements of the Hessian) using finite differences: \begin{equation}
		\frac{\partial^2 L(w)}{\partial w_i^2} \approx \frac{1}{r} \left( \frac{\partial L(w+r)}{\partial w_i} - \frac{\partial L(w)}{\partial w_i} \right)
	\end{equation}
	
	\item Construct the diagonal Hessian matrix $H_d$ using the estimated second-order partial derivatives.
	
\end{itemize}

It was noted that the quality of this approximation depends on the specific problem and the choice of $r$. A small r value may introduce numerical errors, while a large $\epsilon$ value may produce inaccurate approximations. Also, the diagonal approximation assumes that the off-diagonal elements of the Hessian are small, which may not be true for all problems. In such cases, empirically $10^{-07}$ was found to be the most adequate in convolutional deep learning models.

Once the diagonal element of the hessian matrix are obtained. the weights will be updated based on newton method. The update rule for Newton's method involves multiplying the inverse of the Hessian matrix by the gradient of the loss function to obtain the weight update. However, the step size determined by the inverse Hessian can sometimes be too large, leading to unstable convergence. To address this, a learning rate is introduced to scale the weight update. The update rule becomes:
\vspace{1em}

\begin{equation}
	w_{\text{new}} = w_{\text{old}} - r \cdot H^{-1} \cdot \frac{\partial L(w)}{\partial w_i}
\end{equation}
\vspace{1em}

Where $H$ is a matrix contains only the approximated diagonal element. The inverse of this matrix is trivial. $r$ is the learning rate which is a small positive number that determines the step size in the weight update. The learning rate $r$  can be set manually or can be adapted during training using algorithms such as Adagrad or Adam. For our method the learning rate will be capped under arbitrary max value to inhibit the algorithm from overshot if the hessian matrix diagonal element are too small.

\begin{itemize}
	\item Initialize the weight vector \(w\).
	\item Compute the gradient of the loss function with respect to the weights with back propagation.
	\item Advance the algorithm one step with a small learning rate \(r\) and compute again the gradient of the loss function with respect to the weights with back propagation.
	\item Compute the diagonal approximation of the Hessian matrix of the loss function with respect \(H_d\) using equation (6).
	\item Add a positive small number \(\epsilon\) to \(H_d\) to overcome dividing by zero when inverting the \(H_d\) matrix.
	\item Compute the inverse of the \(H_d\) matrix.
	\item Consider only the positive values of the matrix \(H_d\) to not be attracted to saddle points in the gradient descent stepping forward.
	\item Compute the weight update using the update rule \(w_\text{new} = w_\text{old} - \text{learning\_rate} \cdot (H_\text{d})^{-1} \cdot \text{gradient}\).
	\item Limit the weight by an arbitrary max value to not overshoot.
	\item Update the weight vector \(w\) to \(w_\text{new}\).
	\item Until convergence
\end{itemize}

\begin{algorithm}
	\caption{Second order Adaptive Learning Optimizater: SALO}
	\begin{algorithmic}[1]

					\State \textbf{Required parameters:}
					\begin{itemize}
						\item Learning rate $\alpha$ (typically around 0.001)
						\item $\beta_1$ (typically 0.95)
						\item $\beta_2$ (typically 0.97)
						\item $\beta_3$ (typically 0.9)
						\item $\epsilon$ (small constant, typically $1\times10^{-8}$)
						\item Time step $t$ initialized to 0
					\end{itemize}
					\State \textbf{Initialize} each parameter:
					\begin{itemize}
						\item $m_t = 0$
						\item $v_t = 0$
						\item $\hat{m}_t = 0.001$
						\item $\hat{v}_t = 0.001$
						\item Hessian $hess = 0.01$
						\item Learning step $\Delta w = 0.01$
						\item Second derivative $H = 0.001$
						\item Second moment estimate $\hat{H} = 0.001$
						\item Previous gradient $P_G = 0.0$
						\item Initialize $\epsilon$ $eps = 1\times10^{-8}$
					\end{itemize}
					\For{each epoch}
					\For{each mini-batch of data}
					\State Compute gradient of loss function: $g_t = \nabla_\theta(loss)$
					\State Update first moment estimate: $m_t = \beta_1 m_t + (1 - \beta_1)g_t$
					\State Update second moment estimate: $v_t = \beta_2 v_t + (1 - \beta_2)g_t^2$
					\State Estimate second derivative: $der2nd = \frac{P_G - m_t}{\Delta w + eps}$
					\State Prepare gradient for next step: $p_G = m_t$
					\State Update third moment: $H = \beta_3 H + (1 - \beta_3)der2nd$
					\State Correct bias in first moment estimate: $\hat{m}_t = \frac{m_t}{1 - \beta_1^t}$
					\State Correct bias in second moment estimate: $\hat{v}_t = \frac{v_t}{1 - \beta_2^t}$
					\State Update second derivative moment estimate: $\hat{H} = \frac{H}{1 - \beta_3^t}$
					\State Take absolute value of $\hat{H}$ : $\hat{H} = \lvert\hat{H}\rvert$
					\State Clip $\hat{H}$ to prevent overshooting: $\hat{H}(\hat{H} \le 0.01) = 0.01$
					\State Update parameters: $\theta = \theta - \alpha \frac{\hat{m}_t}{(\hat{H} + \epsilon)(\sqrt{\hat{v}_t} + \epsilon)}$
					\EndFor
					\EndFor
					

	\end{algorithmic}
\end{algorithm}

\begin{figure}[ht]
	\centering
	\includegraphics[width=\linewidth]{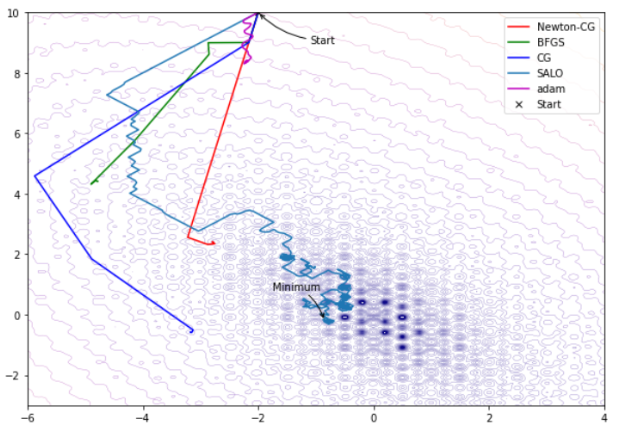}
	\caption{Comparison of our optimizer against ADAM, NewtonCG, BFGS and CG}
	\label{fig:ourmeth}
\end{figure}

We compared our approximation with other methods such as “Newton”, “BFGS”, “CG” and “ADAM” on a generic function, the result is shown in figure \ref{fig:ourmeth}. The new method is apparently less susceptible to local minima.

\subsection{Initial Experimental Results: Discussion On Enhancing the Method}
The off-Hessian method approximates the Hessian matrix by calculating its diagonal elements using finite difference. This approach reduces the computational complexity associated with calculating the full Hessian matrix, enabling the method to be more scalable for large deep learning problems. To address the noisy landscape of the loss function, our off-Hessian method besides a momentum of the first gradient introduces a momentum term for the second derivative, inspired by Adam algorithm (Kingma and Ba, 2014), The momentum assists the optimization algorithm in navigating through noisy regions, facilitating a more effective search for the optimal solution. It was noticed that incorporating second momentum for the second order derivative gives the algorithm a boost to  overcome small local minima so overall the algorithm can reach higher accuracies comparing to Adam.

\section{Fuzzy Logic Based Scheduling: Our Implementation}\label{sec6}

In this research, we propose a fuzzy logic-based scheduler to dynamically adjust both the learning rate and the second derivative momentum during the training process of machine learning models. The scheduler considers two input variables and generates two output variables based on a set of rules. The advantage of using Fuzzy logic system is that we don’t need to know exactly the decay function. It is enough to build the estimator based on our knowledge of the domain such that the learning rate should be smaller at the end of the iteration and should be bigger if the lost function is big and smaller if the loss function is small.
So, the input variable will be as follows:

\begin{itemize}
	\item Loss Behavior: A continuous variable ranging from 0 to 6, partitioned into four fuzzy sets: Best (0-0.6), Acceptable (0.6-1), Medium (1-3), and Worst (3-6).
	
	\begin{figure}
		\centering
		\begin{tabular}{@{}c@{\hspace{2mm}}c@{}}
			\includegraphics[width=0.45\columnwidth]{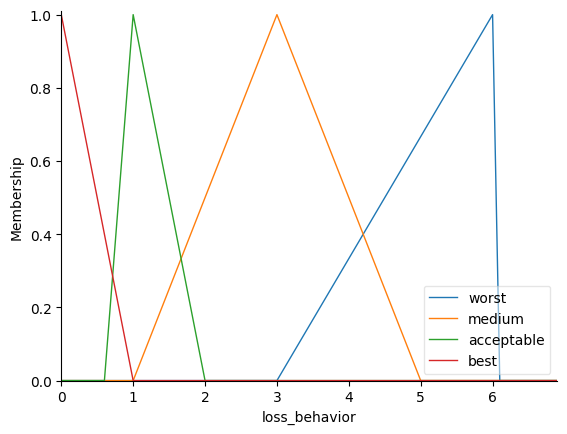} &
			\includegraphics[width=0.45\columnwidth]{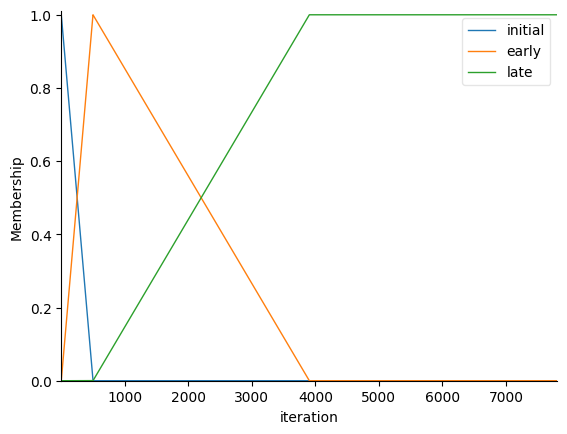}  \\
			(a) Loss Behavior & (b) Iteration
		\end{tabular}
		\caption{Membership Functions of Fuzzy System Inputs}
		\label{fig:mmbr}
	\end{figure}

	\item Training Iteration: A discrete variable ranging from 1 to the total number of training steps, partitioned into three fuzzy sets: Initial (1 to 110), Early (100 to midpoint), and Late (100 to 3/4 of the total steps).
	
\end{itemize}

The output variables are:

\begin{itemize}
	\item Learning Rate: A continuous variable in the range of 0 to $1.5 \times 10^{-7}$, divided into three fuzzy sets: Minimum $(0 $ to $0.5 \times 10^{-7})$, Medium $(0$ to $1 \times 10^{-7})$, and High $(0.5 \times 10^{-7}$ to $1 \times 10^{-7})$.
	
	\item Second Moment Decay Rate $\beta_1$: A continuous variable defined using three fuzzy sets: Minimum (range [0, 0.1, 0.3]), Medium (range [0.1, 0.6, 0.9]), and Maximum (range [0.6, 0.9, 0.99]).
	
	\item Third Moment Decay Rate $\beta_3$: A continuous variable defined using three fuzzy sets: Minimum (range [0, 0.1, 0.3]), Medium (range [0.1, 0.6, 0.9]), and Maximum (range [0.6, 0.9, 0.99]).
	
\end{itemize}

\begin{figure}
	\centering
	\captionsetup{justification=centering}
	\includegraphics[width=\linewidth]{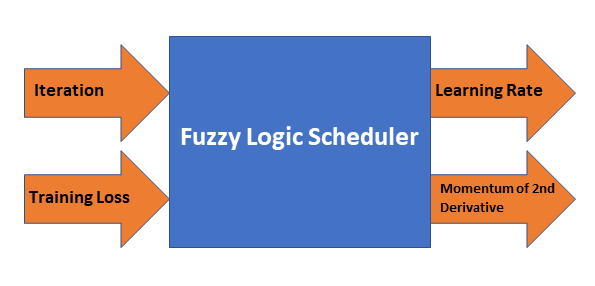}
	\caption{Bloc Diagram of Our Fuzzy Scheduling System}
	\label{fig:diagram}
\end{figure}

The fuzzy scheduler is governed by a set of rules that capture the relationship between the input variables and the output variables. These rules are defined based on expert knowledge and intuition and are meant to guide the training process by dynamically adapting the learning rate, $\beta_1$ and $\beta_3$ according to the model's loss behavior and the current training iteration. By employing this fuzzy logic scheduler, we aim to optimize the learning rate and the second derivative momentum during training, potentially improving model performance and convergence speed.

A control system is constructed using these rules, and a control system simulation object is initialized for making predictions. The rules governing the relationship between the input variables and the output variables are designed based on expert knowledge and intuition to optimize the training process further. The inclusion of $\beta_1$ and $\beta_3$ adds another dimension to the control system, allowing for more fine-tuned adjustments during the training process.

\begin{figure}
	\centering
	\begin{tabular}{@{}c@{\hspace{2mm}}c@{\hspace{2mm}}c@{}}
		\includegraphics[width=0.3\columnwidth]{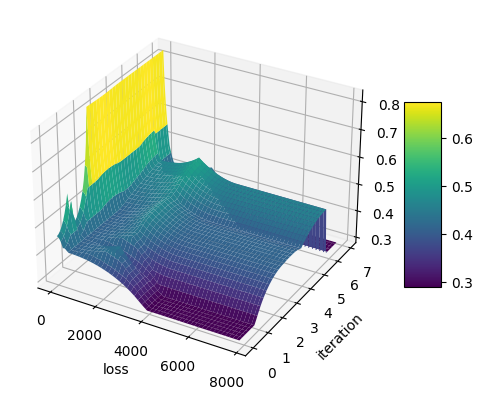} &
		\includegraphics[width=0.3\columnwidth]{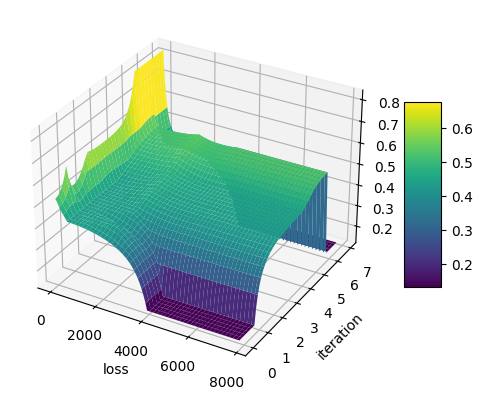} &
		\includegraphics[width=0.3\columnwidth]{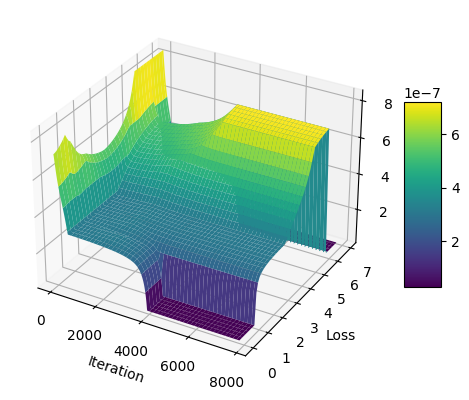} \\
		(a) $\beta_1$ Variation & (b) $\beta_3$ Variation & (c) Learning Rate
	\end{tabular}
	\caption{Output Variables}
	\label{fig:lrmmntm}
\end{figure}

\section{Empirical Performance: Application on ImageNet}\label{sec8}

We have tested our optimisation method, refered to as SALO, on the popular ImageNET dataset, ImageNet is a large visual database designed for use in visual object recognition research. It is often used as a benchmark for image classification algorithms. 

\subsection{Experiment Setup}\label{sec:nofls}

Our experiments were conducted using Google Colab Premium Pro, which provided a robust environment with 83.5 GB of RAM and a 52 GB NVIDIA P100 GPU. The training process consisted of 5 epochs for all optimization methods.

In our experiments, we employed the EfficientNet-B3 architecture as the backbone model for our tasks. EfficientNet-B3 is a state-of-the-art convolutional neural network (CNN) that is part of the EfficientNet family of models, introduced by Tan and Le (2019) \citep{Tan2019}. The core idea behind EfficientNet is to perform model scaling in a more structured manner, using a compound coefficient, which simultaneously scales the network depth, width, and resolution. This approach, called compound scaling, allows the model to maintain a balance between these three dimensions, resulting in better accuracy and efficiency trade-offs.

EfficientNet-B3 is the third model in the EfficientNet series and has a higher capacity compared to the B0 and B1 variants. It consists of approximately 12 million parameters and has a depth of 154 layers. The model incorporates depthwise separable convolutions, residual connections, and squeeze-and-excitation modules to reduce computational complexity while maintaining high performance. It has been pre-trained on the ImageNet dataset and has demonstrated competitive performance in various computer vision tasks, such as object recognition, segmentation, and fine-grained classification.

We chose the EfficientNet-B3 architecture because it offers a suitable balance between model complexity and computational efficiency, making it a suitable choice for our experiments conducted on Google Colab Premium Pro. Using EfficientNet-B3 allows us to leverage a well-established, high-performance model, ensuring a reliable foundation for our optimization experiments with SALO and other optimizer algorithms.

We compare SALO against well-established optimizers such as SGD, Adam, and AdamW. To establish a robust baseline, we employ the optimal hyperparameters documented in the literature for SGD, Adam, and AdamW. We perform only minimal tuning on SALO for two reasons: first, we lack access to industrial-scale resources for extensive tuning, and second, our objective is to demonstrate SALO's average performance rather than its best possible performance achieved through exhaustive tuning. Consequently, we directly apply the same $\mathrm{\beta}_1$, $\mathrm{\beta}_2$, weight decay, batch size, and learning rate schedule from Adam to SALO, although adjusting these parameters could potentially enhance SALO's performance. For SALO, we focus on tuning the learning rate, second and third derivative momentum.

\subsection{Experimental Results: TinyImageNet} \label{sec:fls}

By the end of the training, SALO achieved a substantially lower training loss of 0.068 and higher validation accuracy of 0.820 compared to Adam, which resulted in a training loss of 0.437 and validation accuracy of 0.766. Additionally, SALO outperformed SGD with a training loss of 0.390 and validation accuracy of 0.725, and AdamW with a training loss of 0.609 and validation accuracy of 0.665. This significant reduction in training loss and increase in validation accuracy demonstrate the effectiveness and potential of our SALO optimization method in comparison to well-established optimizers such as Adam, SGD, and AdamW.

\begin{table}[ht]
	\setlength{\abovecaptionskip}{8pt} 
	\setlength{\belowcaptionskip}{0pt}
	\centering
	\begin{tabularx}{\columnwidth}{@{}l*{6}{X}@{}}
		\toprule
		Method & Epoch & Epoch Loss & Train Loss & Train Acc. & Val. Loss & Val. Acc. \\
		\midrule
		\multirow{5}{*}{Adam}
		& 1     & 1.42       & 1.160    & 0.699             & 1.513    & 0.631             \\
		& 2     & 0.67       & 0.481    & 0.868             & 0.977    & 0.754             \\
		& 3     & 0.56       & 0.440    & 0.879             & 0.955    & 0.763             \\
		& 4     & 0.37       & 0.437    & 0.880             & 0.958    & 0.763             \\
		& 5     & 0.60       & 0.437    & 0.880             & 0.954    & 0.766             \\
		\midrule
		\multirow{5}{*}{SALO}
		& 1     & 1.32       & 0.536    & 0.854             & 0.905    & 0.767             \\
		& 2     & 0.70       & 0.335    & 0.905             & 0.912    & 0.772             \\
		& 3     & 0.21       & 0.133    & 0.963             & 0.779    & 0.811             \\
		& 4     & 0.09       & 0.093    & 0.975             & 0.765    & 0.817             \\
		& 5     & \textbf{0.26}       & \textbf{0.068}    & \textbf{0.982}             & \textbf{0.769}    & \textbf{0.820}             \\
		\midrule
		\multirow{5}{*}{SGD}
		& 1     & 1.55       & 1.025    & 0.743             & 1.277    & 0.687             \\
		& 2     & 0.81       & 0.394    & 0.890             & 0.832    & 0.789             \\
		& 3     & 0.18       & 0.295    & 0.916             & 0.868    & 0.780             \\
		& 4     & 0.46       & 0.174    & 0.950             & 0.831    & 0.795             \\
		& 5     & 0.54       & 0.390    & 0.891             & 1.204    & 0.725             \\
		\midrule
		\multirow{4}{*}{AdamW}
		& 1     & 2.10       & 1.253    & 0.680             & 1.597    & 0.608             \\
		& 2     & 1.40       & 0.960    & 0.745             & 1.482    & 0.638             \\
		& 3     & 0.92       & 0.792    & 0.784             & 1.464    & 0.649             \\
		& 4     & 1.56       & 0.609    & 0.830             & 1.442    & 0.665             \\
		\bottomrule
	\end{tabularx}
	\caption{Comparison of SALO against Adam, AdamW and SGD. Best scores highlighted in bold.}
	\label{tab:comparison}
\end{table}

\begin{figure}
	\centering
	\captionsetup{justification=centering}
	\includegraphics[width=\linewidth]{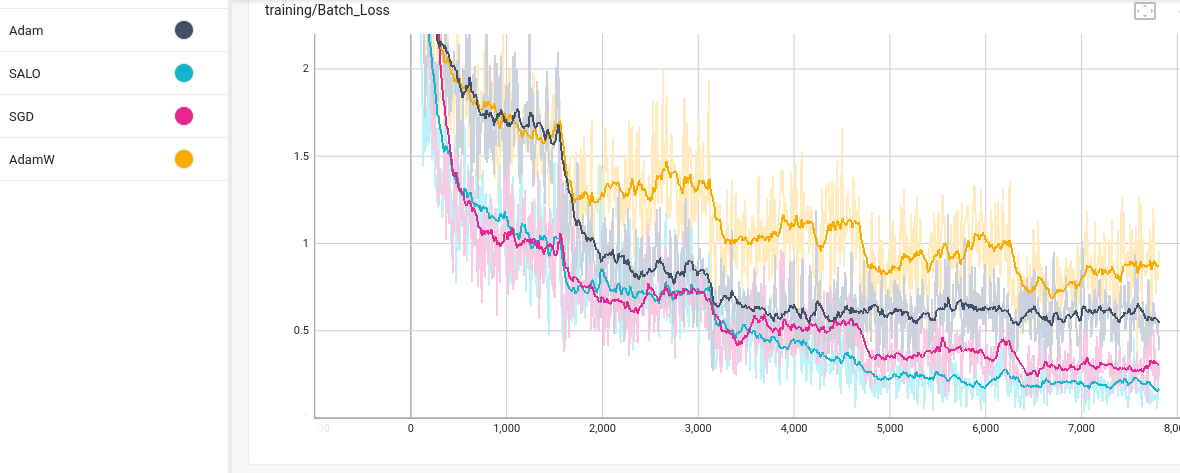}
	\caption{Training Accuracy: SALO VS. Adam , AdamW and SGD}
	\label{fig:diagram}
\end{figure}

\begin{figure}
	\centering
	\captionsetup{justification=centering}
	\includegraphics[width=\linewidth]{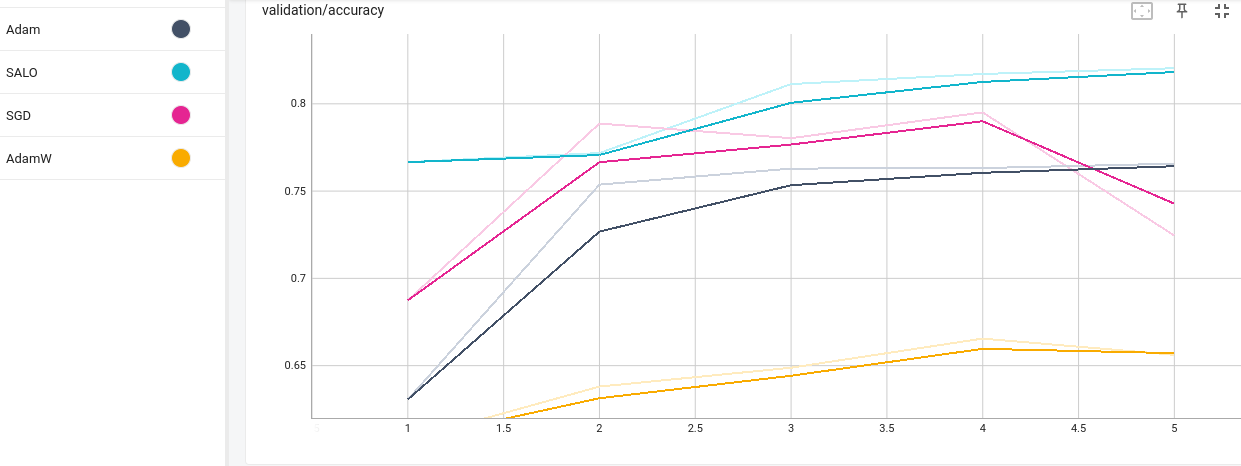}
	\caption{Validation Accuracy: SALO VS. Adam , AdamW and SGD}
	\label{fig:diagram}
\end{figure}

\subsection{Experimental Results: Discussion} \label{sec:expdsic}

We would like to highlight that the calculation of fuzzy parameters in the proposed SALO optimizer takes approximately 20 minutes longer than other methods such as Adam,SGD, and AdamW butit reaches -with less iterations- a better accuracy even for the same time. Although employing a lookup table from the fuzzy system for fuzzy parameters could significantly reduce this computation time, we opted for the direct calculation of the fuzzy system due to its flexibility in the experimental stage. The lookup table can be calculated from the fuzzy system one time in the beginning of the optimization. This will allow faster computation, as precomputed fuzzy parameters can be rapidly accessed and applied during training. This lowers runtime overhead, as the complexity of fuzzy calculations is shifted to the pre-computation phase. When we conducted our experiments on the TinyImageNet dataset with “EfficientNet 3b” which is relatively small (12 million) compared to the some of the largest models in use today the additional computation time was not a significant concern. For bigger models we suggest to prepare a lookup table ahead before starting the optimization. In this paper we didn’t explore the full fuzzy rules space. We applied rules based on general knowledge and expertise on the matter. It is worth further investigating the impact of the different rules ensemble on the computation as well as exploring different combinations of first and second derivatives momentum values. The hyperparameters presented in this paper should be considered as proof of concept, demonstrating the potential of the Second order Adaptive Learning Optimizer SALO, rather than the definitive best values. In future work, leveraging more powerful computational resources and more extensive hyperparameters search techniques could lead to further improvements in the performance of the SALO optimizer

\section{Conclusion and Future Work}\label{sec9}

In our study, we developed a second-order optimization algorithm for deep learning that effectively minimizes computational burden while achieving robust performance. By implementing an online calculation of the diagonal Hessian matrix at each step, the algorithm's resource requirements are comparable to those of the widely-used Adam optimizer. To further improve second derivative approximation and address variations and the uniformity in learning rates, our method incorporates first derivative momentum, which reduces noise in the calculations similarly to the Adam algorithm. Additionally, our approach utilizes the momentum of the second derivative to ensure a smoother descent, offering increased resistance to noisy landscapes. Empirical results indicate that a starting learning rate of 1/1000000 is optimal. We found that lower learning rates lead to smaller errors in estimating the second derivatives in the diagonal Hessian matrix.  Newton method optimization is fundamentally different than gradient descent. When the second derivative is zero or very small the rate of change shoots to infinity.  in our algorithm the second derivative is allowed to be as minimum as an empirical value 0.0001 and not less to overcome the shoot to infinity. our algorithm overcomes this by setting an empirical minimum value of 0.0001 for the second derivative, preventing the rate of change from shooting to infinity. Lastly, we recommend optimizing the algorithm by running it on a smaller subset of the dataset to identify the best parameters before applying it to the full dataset, thus ensuring better overall performance.

\bibliographystyle{unsrtnat}
\bibliography{references}  
\end{document}